\pdfoutput=1

\documentclass[11pt]{article}

\usepackage[]{acl}

\usepackage{algorithm}
\usepackage{algorithmic}

\usepackage{times} 
\usepackage{latexsym}

\usepackage[T1]{fontenc}

\usepackage[utf8]{inputenc}

\usepackage{microtype}

\usepackage{graphicx}
\usepackage{amsmath}
\usepackage{multirow}
\usepackage{comment}
\usepackage{booktabs}
\usepackage{multirow}

\newcommand{\cut}[1]{}

\title{Are Current Decoding Strategies Capable of Facing\\ the Challenges of Visual Dialogue?}

\author{\stepcounter{footnote}Amit Kumar Chaudhary\\
  CIMeC, University of Trento \\
  \texttt{amitkumar.chaudhar@unitn.it} \\\And
    Alex J. Lucassen \\
  CIMeC, University of Trento \\
  \texttt{alex.lucassen@unitn.it} \\\AND
  Ioanna Tsani \\
  CIMeC, University of Trento \\
  \texttt{ioanna.tsani@unitn.it} \\\And
  Alberto Testoni\\
  DISI, University of Trento \\
  \texttt{alberto.testoni@unitn.it} 
  }

\begin{document}
\maketitle
\begin{abstract}

Decoding strategies play a crucial role in natural language generation systems. They are usually designed and evaluated in open-ended text-only tasks, and it is not clear how different strategies handle the numerous challenges that goal-oriented multimodal systems face (such as grounding and informativeness). To answer this question, we compare a wide variety of different decoding strategies and hyper-parameter configurations in a Visual Dialogue referential game. Although none of them successfully balance lexical richness, accuracy in the task, and visual grounding, our in-depth analysis allows us to highlight the strengths and weaknesses of each decoding strategy. We believe our findings and suggestions may serve as a starting point for designing more effective decoding algorithms that handle the challenges of Visual Dialogue tasks.




\end{abstract}

\section{Introduction}
\label{sec:introduction}








The last few years have witnessed remarkable progress in developing efficient generative language models. The choice of the decoding strategy plays a crucial role in the quality of the output (see \citet{decoding_survey} for an exhaustive overview). It should be noted that decoding strategies are usually designed for and evaluated in text-only settings. The most-used decoding strategies can be grouped into two main classes. On the one hand, decoding strategies that aim to generate text that maximizes likelihood (like greedy and beam search) are shown to generate generic, repetitive, and \textit{degenerate} output. \citet{zhang-etal-2021-trading} refer to this phenomenon as \textit{the likelihood trap}, and provide evidence that these strategies lead to sub-optimal sequences. On the other hand, stochastic strategies like pure sampling, top-k sampling, and nucleus sampling \cite{holt:thecu20} increase the variability of generated texts by taking random samples from the model. However, this comes at the cost of generating words that are not semantically appropriate for the context in which they appear. Recently, \citet{typical_paper} used an information-theoretic framework to propose a new decoding algorithm (typical decoding), which samples tokens with an information content close to their conditional entropy. Typical decoding shows promising results in human evaluation experiments but, given its recent release, it is not clear yet how general this approach is. 



\begin{figure}[t]
    \includegraphics[width=1\linewidth]{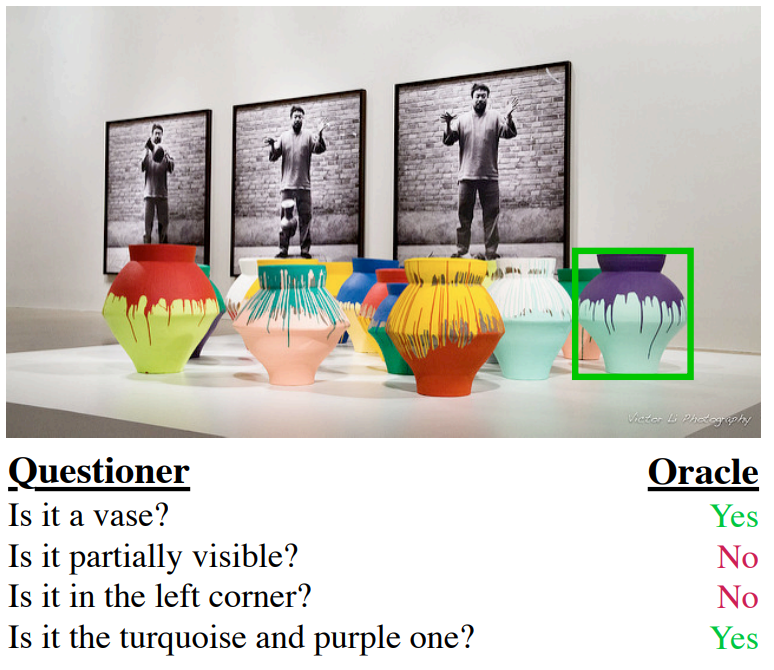}
    \centering
    \caption{Example of a GuessWhat game from \citet{guesswhat_game}}
    \label{fig:gw_ex}
\end{figure}

Multimodal vision \& language systems have recently received a lot of attention from the research community, but a thorough analysis of different decoding strategies in these systems has not been carried out. Thus, the question arises of whether the above-mentioned decoding strategies can handle the challenges of multimodal systems. i.e., generate text that not only takes into account lexical variability, but also grounding in the visual modality. Moreover, in goal-oriented tasks, the informativeness of the generated text plays a crucial role as well. To address these research questions, in this paper we take a referential visual dialogue task, GuessWhat?!\@ \cite{guesswhat_game}, where two players (a Questioner and an Oracle) interact so that the Questioner identifies the secret object assigned to the Oracle among the ones appearing in an image (see Figure \ref{fig:gw_ex} for an example). Apart from well-known issues, such as repetitions in the output, this task poses specific challenges for evaluating decoding techniques compared to previous work. On the one hand, the generated output has to be coherent with the visual input upon which the conversation takes place. As highlighted by \citet{rohrbach-etal-2018-object, testoni-bernardi-2021-ive}, multimodal generative models often generate \textit{hallucinated} entities, i.e., tokens that refer to entities that do not appear in the image upon which the conversation takes place. On the other hand, the questions must be informative, i.e., they must help the Questioner to incrementally identify the target object.



We show that the choice of the decoding strategy and its hyper-parameter configuration heavily affects the quality of the generated output. Our results highlight the specific strengths and weaknesses of decoding strategies that aim at generating sequences with the highest probability vs. strategies that randomly sample words. We find that none of the decoding strategies currently available is able to balance task accuracy and linguistic quality of the output. However, we also show which strategies perform better at important challenges, such as incremental dialogue history, human evaluation, hallucination rate, and lexical diversity. We believe our work may serve as a starting point for designing decoding strategies that take into account all the challenges involved in Visual Dialogue tasks.

\section{Task \& Dataset}
\label{sec:task_data}

GuessWhat?!\@ \cite{guesswhat_game} is a simple object identification game in English where two participants see a real-world image from MSCOCO \cite{lin:2014} containing multiple objects. One player (the Oracle) is secretly assigned one object in the image (the target) and the other player (the Questioner) has to guess it by asking a series of binary yes-no questions to the Oracle. The task is considered to be successful if the Questioner identifies the target. The dataset for this task was collected from human players via Amazon Mechanical Turk. The authors collected 150K dialogues with an average of 5.3 binary questions per dialogue. Figure \ref{fig:gw_ex} shows an example of a GuessWhat game from the dataset.

\section{Model and Decoding Strategies}
\label{sec:model_strategies}
We use the model and pre-trained checkpoints of the Questioner agent made available by \citet{testoni2021looking} for the GuessWhat?!\@ task. This model is based on the GDSE architecture \cite{shekhar-etal-2019-beyond}. It uses a ResNet-152 network \cite{he2016:resnet} to encode the images and an LSTM network to encode the dialogue history. A multimodal shared representation is generated and then used to train both the question generator (which generates a follow-up question given the dialogue history) and the Guesser module (which selects the target object among a list of candidates at the end of the dialogue) in a joint multi-task learning fashion. \citet{testoni2021looking} added an internal Oracle module to the GDSE architecture, which guides a cognitively-inspired beam search re-ranking strategy (\textit{Confirm-it}) at inference time: this strategy promotes the generation of questions that aim at confirming the model's intermediate conjectures about the target. In our work, at inference time the Questioner agent always interacts with the baseline Oracle agent proposed in \citet{guesswhat_game}.

We analyse the effect of a large number of decoding strategies as well as hyper-parameter configuration for each strategy: as highlighted by \citet{zhang-etal-2021-trading}, it is crucial to evaluate different hyper-parameter configurations when comparing multiple decoding strategies. Among the ones that maximize the likelihood of the sequence, we consider plain \textbf{beam search} (with a beam size of 3) and \textbf{greedy search}. We also consider \textbf{Confirm-it}, the cognitively-inspired beam search re-ranking strategy proposed in \citet{testoni2021looking} for promoting the generation of questions that aim at confirming the  model's intermediate conjectures about the target. This strategy re-ranks the set of candidate questions from beam search and selects the one that helps the most in confirming the model's hypothesis about the target. As for stochastic strategies, we analyse \textbf{pure sampling}, \textbf{top-k sampling} (with different $k$ values), and \textbf{nucleus sampling} (with different $p$ values), a strategy proposed in \citet{holt:thecu20} which selects the highest probability tokens whose cumulative probability mass exceeds a given threshold $p$. We also consider \textbf{typical decoding} (with different $\tau$ values), a recently proposed strategy \cite{typical_paper} based on an information-theoretic framework. We refer to the respective papers for additional details on decoding strategies. We let the model generate 5 questions\footnote{Except for the accuracy per turn metric in Section \ref{sec:turn_acc}, where the dialogues consisted of 10 questions.} at test time and average the results over five random seeds.




\section{Metrics}
\label{sec:metrics}

We are interested in evaluating different decoding strategies against a set of metrics that reflect the complexity of the different skills required to successfully solve multimodal referential games. 

\textbf{Linguistic Quality:} We compute the percentage of games with at least one repeated question, the overall number of unique words used by the model and, in line with the observations in \citet{testoni2021interplay}, the number of \textit{rare words} generated by the model, defined as those words that appear fewer than 20 times in the training set. 

\textbf{Visual Grounding:} To quantify the rate of object hallucination in the generated dialogues, we compute the CHAIR metric \cite{rohrbach-etal-2018-object, testoni-bernardi-2021-ive}. This metric, originally proposed for image captioning, detects hallucination by checking each object mentioned in a generated image caption against the ground-truth MSCOCO objects for that image. The metric consists of two distinct variants: CHAIR-i, or per-instance variant (number of hallucinated objects divided by the total number of objects mentioned in each dialogue), and CHAIR-s, or per-sentence variant (number of dialogues with at least one hallucination divided by the total number of dialogues).\footnote{\citet{testoni-bernardi-2021-ive} first adapted the CHAIR metric for Visual Dialogue. However, the authors did not investigate the effect of different decoding strategies.}

\textbf{Informativeness:} To study the informativeness of the generated questions, we report the raw accuracy of the model in guessing the target object after each dialogue turn and at the end of the dialogue. A game is considered successful if the model identifies the target object assigned to the Oracle. Similarly, we also report the accuracy of human annotators when guessing the target by reading machine-generated dialogues. 



\section{Results}
\label{sec:results}

\begin{table*}[]
	\centering
	\fontsize{8}{8}\selectfont
	\bgroup
    \def\arraystretch{1.5}
	\begin{tabular}{|c|c|c|c|c|c|c|}
		\hline
		& \begin{tabular}[c]{@{}c@{}}Accuracy (\%) $\uparrow$ \end{tabular} & CHAIR-i $\downarrow$ & CHAIR-s $\downarrow$ & \begin{tabular}[c]{@{}c@{}}\% games with\\ repetitions $\downarrow$ \end{tabular}  & \begin{tabular}[c]{@{}c@{}}Vocabulary\\ Size $\uparrow$ \end{tabular} &  \begin{tabular}[c]{@{}c@{}}Rare\\ Words $\uparrow$ \end{tabular}  \\ \hline
		Confirm-it & \textbf{51.39} & \textbf{15.09} & \textbf{28.48} & 30.33 & 858 & 34\\ \hline
		Beam Search (beam size = 3) & \textbf{47.05} & 18.33 & \textbf{31.08} & 38.49 & 731 & 27\\ \hline
		Nucleus Sampling ($p$ = 0.3) & \textbf{46.92} & \textbf{17.96} & 33.60 & 32.35 & 1016 & 78\\ \hline
		Greedy Search & 46.58 & \textbf{17.75} & \textbf{32.97} & 35.63 & 834 & 46\\ \hline
		Typical Decoding ($\tau$ = 0.7) & 45.45 & 21.84 & 37.81 & \textbf{16.18} & \textbf{1703} & \textbf{247}\\ \hline
		Top-k Sampling ($k$ = 5) & 45.10 & 22.84 & 37.71 & \textbf{14.93} & \textbf{1462} & \textbf{171}\\ \hline
		Pure Sampling & 43.13 & 26.55 & 43.23 & \textbf{8.32} & \textbf{2609} & \textbf{793}\\ \hline

	\end{tabular}
	\egroup
	\caption{Comparison between decoding strategies and their best-performing (in terms of accuracy) hyper-parameters. The decoding strategies are sorted by accuracy. 
	}
	\label{tab:results}
\end{table*}

\subsection{Quantitative Results} 

Table \ref{tab:results} shows the performance of different decoding strategies against accuracy and dialogue quality, as described by the metrics in Section \ref{sec:metrics}. \footnote{Here we only report the best-performing configuration for each decoding strategy (see SM for all configurations).}
Confirm-it is by far the best decoding strategy in terms of accuracy and hallucination rate. However, it uses a restricted vocabulary compared to other strategies. A similar issue is observed for greedy and beam search. We find nucleus sampling (with a $p$-value of 0.3, much lower than the one used by the authors in \citet{holt:thecu20}) to effectively increase the lexical variety compared to beam search, without damaging accuracy and hallucination rate. Typical decoding, top-k and pure sampling, instead, clearly decrease repetitions and increase the vocabulary richness by generating tokens that are not related to the source input, as indicated by the high hallucination rate. It thus looks like there exists a trade-off between informativeness / visual grounding and linguistic quality.


\subsection{Effect of Hyper-Parameter Choice}

\begin{figure}[h]
    \includegraphics[width=1\linewidth]{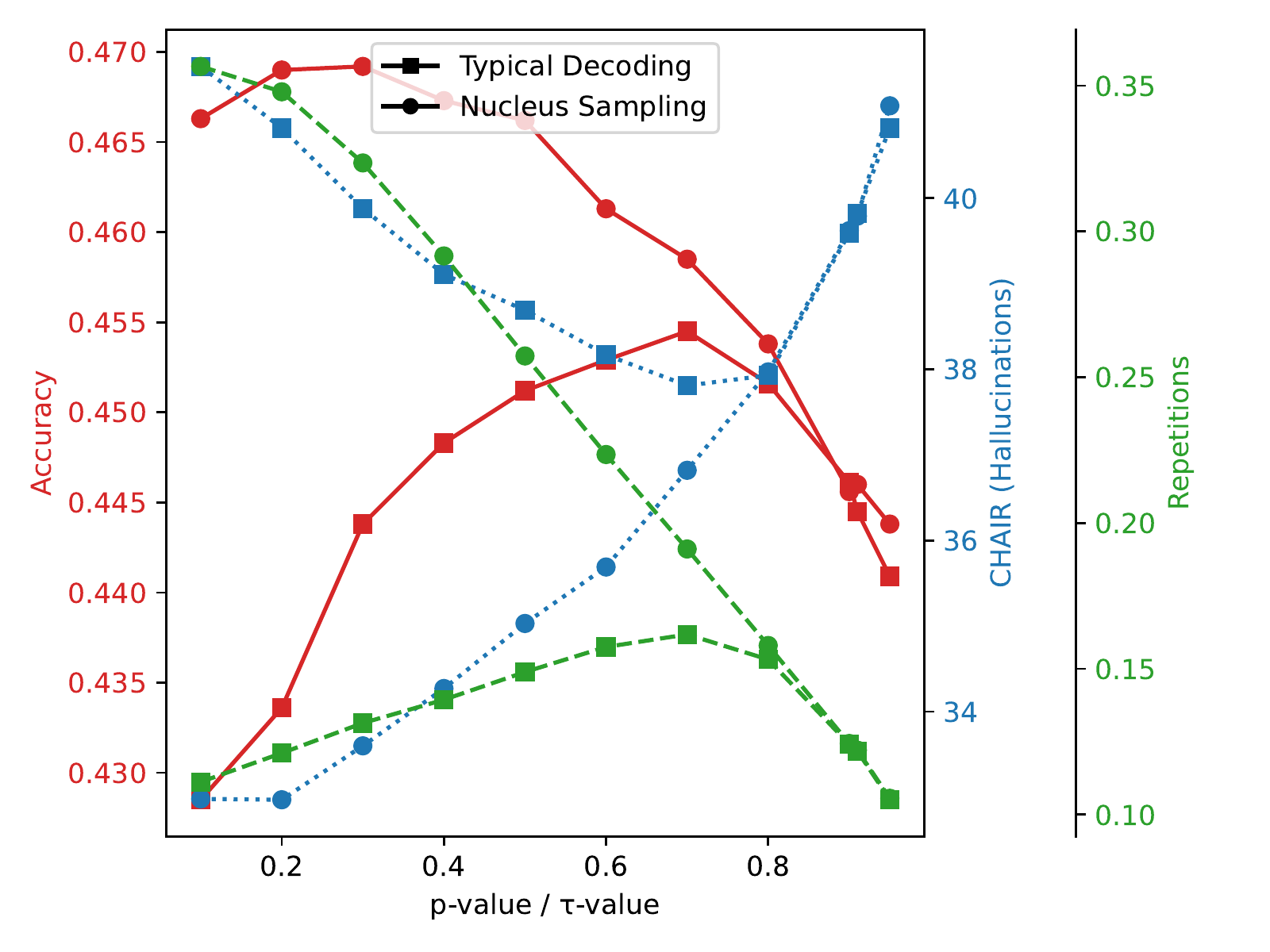} 
    \centering
    \caption{Different hyper-parameter values and their effect on the accuracy, hallucinations, and repetitions in typical decoding and nucleus sampling.
    }
    \label{fig:nucl_typ}
\end{figure}


We study the effect of hyper-parameter configurations in stochastic strategies. Specifically, we try various $p$-values for nucleus sampling and $\tau$-values for typical decoding.\footnote{Results for top-k are in SM.} As shown in Figure \ref{fig:nucl_typ}, both typical and nucleus sampling peak in accuracy with the parameter configurations that also lead to the most repetitions and fewest hallucinations. Conversely, both strategies show the lowest accuracy with the highest hallucination rate. These results confirm the detrimental effect of hallucinations on the performance of the model. It is interesting to note the robustness of typical decoding in generating few repetitions regardless of the $\tau$ value. In line with the findings in \citet{zhang-etal-2021-trading}, this analysis confirms the importance of hyper-parameter configurations and the peculiar trade-off between informativeness, repetitions, and visual grounding: so far it has not been possible to find a single configuration that optimizes all three at the same time. 

\cut{
As highlighted by \citet{zhang-etal-2021-trading}, it is important to analyse the effect of hyper-parameter configurations. Specifically, we tried various $p$-values for nucleus sampling and $\tau$-values for typical decoding, as well as different $k$-values for top-k sampling.\footnote{These values and their results can be seen in detail in the SM (Table \ref{tab:results_full}).} This experimentation additionally revealed some interesting patterns both in the accuracy scores and in the number of hallucinations and repetitions. As shown in Figure \ref{fig:nucl_typ}, typical decoding peaked in accuracy at $\tau$ = 0.7. This was also the point at which hallucinations were at their lowest, and repetitions at their highest. Nucleus sampling, instead, reaches the highest accuracy with $p=0.3$, while CHAIR monotonically increases with $p$. Repetitions follow the opposite path of CHAIR for nucleus sampling while, interestingly, for typical decoding the CHAIR scores seem to be less affected by $\tau$. Top-k sampling produces more hallucinations and decreased accuracy as the $k$-value gets higher, similarly to the effect of higher $p$-values in nucleus sampling. The best-performing $k$-value is 5 (details about the performance of top-k sampling can be found in Figure \ref{fig:topk} of the SM). 
}


\subsection{Per-turn Accuracy}
\label{sec:turn_acc}

\begin{figure}[h]
    \includegraphics[width=1\linewidth]{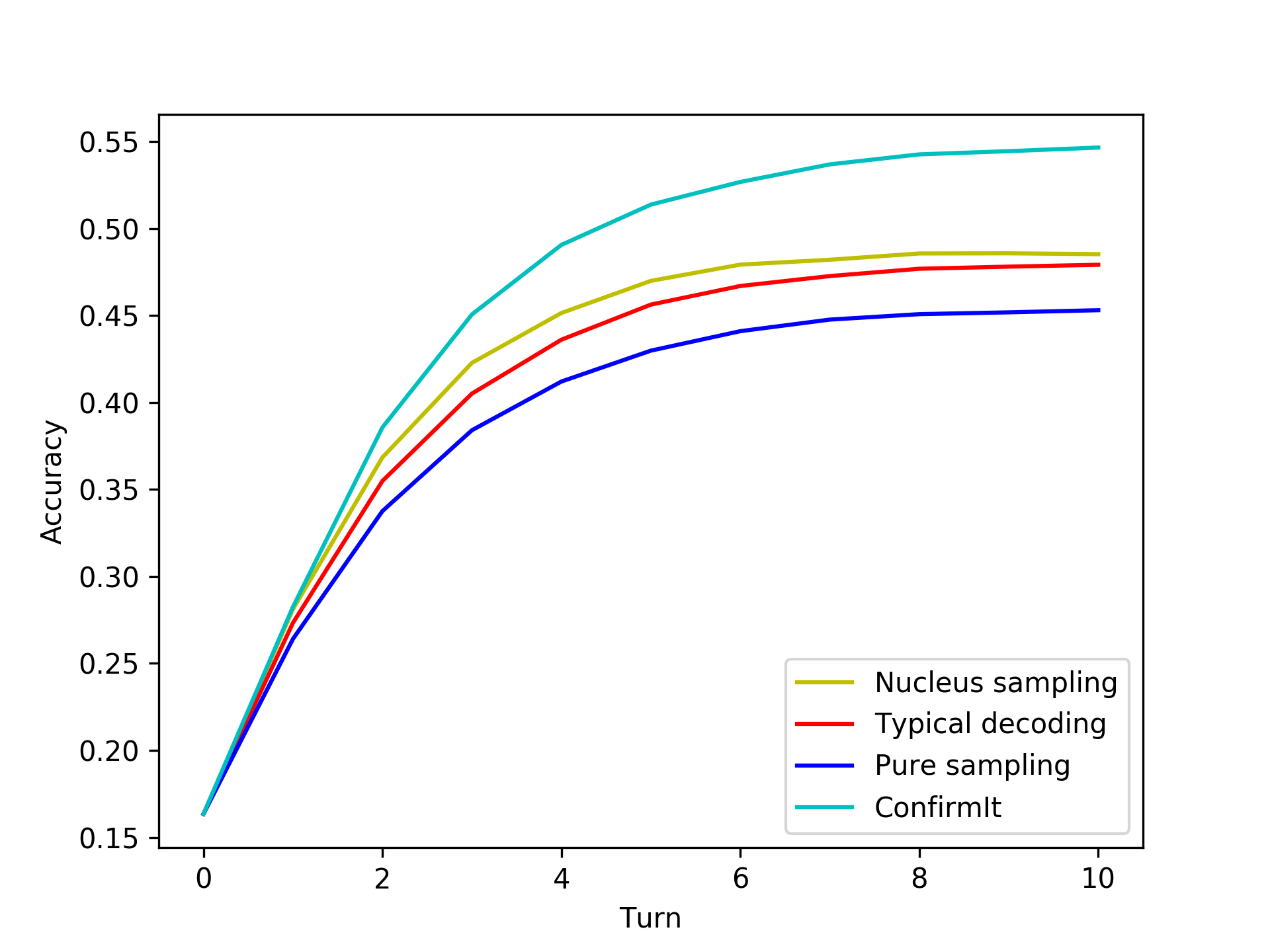} 
    \centering
    \caption{The accuracy per dialogue turn for four different decoding strategies for dialogues of length 10.}
    \label{fig:acc_10turns}
\end{figure}

One crucial ability in GuessWhat?!\@ is asking informative questions that incrementally help in identifying the target: for this reason, we check the accuracy of the model after each new question is asked. Figure \ref{fig:acc_10turns} shows accuracy per dialogue turn for a set of representative strategies: Nucleus sampling ($p$=0.3), Typical Decoding ($\tau$=0.7), Confirm-it, and pure sampling. To get a broader picture, we let the model generate 10 questions in this setting. Confirm-it stands out by showing the largest incremental increase of accuracy throughout the dialogue, indicating that it generates more effective follow-up questions. Pure sampling, on the other hand, seems to suffer from the very beginning of the dialogue and its accuracy stabilizes soon. It is worth noting that the accuracy of typical decoding gets closer to that of nucleus sampling towards the end of the dialogue, with the latter leveling off sooner. We conjecture that Confirm-it outperforms other techniques because it takes into account the probability of the Guesser at inference time, so it is guided to generate questions that change these probabilities and thus avoid generic questions.





\begin{table}[]
	\centering
	\small
	\bgroup
    \def\arraystretch{1.5}
	\begin{tabular}{|c|c|c|c|c|c|c|}
		\hline
		& \begin{tabular}[c]{@{}c@{}}Human Accuracy (\%) $\uparrow$ \end{tabular}\\ \hline
		Confirm-it & \textbf{72.5}\\ \hline
		Typical Sampling ($\tau$ = 0.7) & 68.0\\ \hline
		Nucleus Sampling ($p$ = 0.3) & 67.5\\ \hline
		Pure Sampling & 59.5\\ \hline
	\end{tabular}
	\egroup
	\caption{
	Human Guess Accuracy based on dialogue generated from different decoding strategies.
	}
	\label{tab:human-accuracy}
\end{table}

\subsection{Human Evaluation}
We asked 8 human annotators to guess the target object in a sample of GuessWhat?!\@ games when reading dialogues generated by our model with different decoding strategies. Each participant annotated 100 games (25 per strategy) and the decoding strategy was not revealed during the annotation. As shown in Table \ref{tab:human-accuracy}, humans reach the highest accuracy when reading dialogues generated by Confirm-it, followed by typical decoding and nucleus sampling, while pure sampling falls behind. These results, which do not mirror the accuracy result in Table \ref{tab:results}, allow us to disentangle the weaknesses of the Guesser (i.e., the classification module that predicts the target) from the actual informativeness of the dialogues. Compared to the model, human annotators seem to better exploit the lexical richness of typical decoding and nucleus sampling. We refer to the SM for additional information about the annotation procedure, in line with the best-practice guidelines in \citet{van2021human}.







\section{Related Work}
\label{sec:related}

In the field of multimodal NLG, \citet{zarriess-schlangen-2018-decoding} propose trainable decoding for referring expression generation. The authors propose a two-stage optimization set-up where a small network processes the RNN’s hidden state before passing it to the decoder, using BLEU score as a reward for the decoder. We did not analyse this approach in our paper because we focus only on decoding strategies that do not require any change in the architecture or training of the model. We leave for future work an analysis of trainable decoding approaches. Inspired by the findings in \citet{holt:thecu20}, \citet{massarelli-etal-2020-decoding} propose a hybrid decoding strategy for open-ended text generation which combines the non-repetitive nature of sampling strategies with the consistency of likelihood-based approaches. The authors show that their approach generated less repetitive and more verifiable text. The design of hybrid decoding strategies for multimodal tasks is out of the scope of this paper, but is an interesting subject to pursue in future work.

\section{Discussion and Conclusion}
\label{sec:conclusion}
Decoding algorithms are a key component of natural language generation systems. They are usually designed for and evaluated in text-only tasks. We believe multimodal (vision \& language) and goal-oriented tasks pose unique and under-studied challenges to current decoding strategies. In this paper, we ran an in-depth analysis of several decoding strategies (and their hyper-parameter configurations) for a model playing a referential visual dialogue game.  We found that decoding algorithms that lead to the highest accuracy in the task and the lowest hallucination rate, at the same time generate highly repetitive text and use a restricted vocabulary. Our analyses reveal the crucial role of hyper-parameter configuration in stochastic strategies, an issue that poses several questions about the trade-off between lexical variety, hallucination rate, and task accuracy. While nucleus sampling partially balances the above-mentioned issues, human annotators seem to better exploit the richness of the dialogues generated by typical decoding. Finally, our results demonstrate that a beam search re-ranking algorithm (Confirm-it) generates more effective follow-up questions throughout the dialogue turns. We believe that taking into account the model's intermediate predictions about the referent, like Confirm-it does, represents a promising direction that should be applied also to stochastic strategies in future work, aiming at preserving their lexical richness while reducing hallucinations. 

Our results demonstrate that none of the decoding strategies currently at disposal effectively take into account both task accuracy and dialogue quality at the same time. We also highlight peculiar features of each strategy that may guide future research with the goal of designing decoding strategies that properly confront the crucial challenges of multimodal goal-oriented dialogues. 






\section*{Acknowledgements}

We kindly acknowledge the support of the Department of Information Science and Engineering (DISI) for the arrangement of the GPUs used in our research. We would like to thank Professor Raffaella Bernardi for the suggestions and feedback. The first three authors are enrolled in the Erasmus Mundus European Masters Program in Language and Communication Technologies. The first author is supported by the Erasmus+ Programme. The third author of this scientific paper was supported by the Onassis Foundation - Scholarship ID: F ZR 060-1/2021-2022. We thank the Master's students of the Grounded Language Processing course at the University of Trento for the feedback provided at an early stage of this project.


\bibliography{references}
\bibliographystyle{acl_natbib}

\clearpage

\appendix

\section{Supplementary Material}
\label{sec:supplementary}



\subsection{Effect of Hyper-parameters}
\begin{figure}[h]
    \includegraphics[width=1\linewidth]{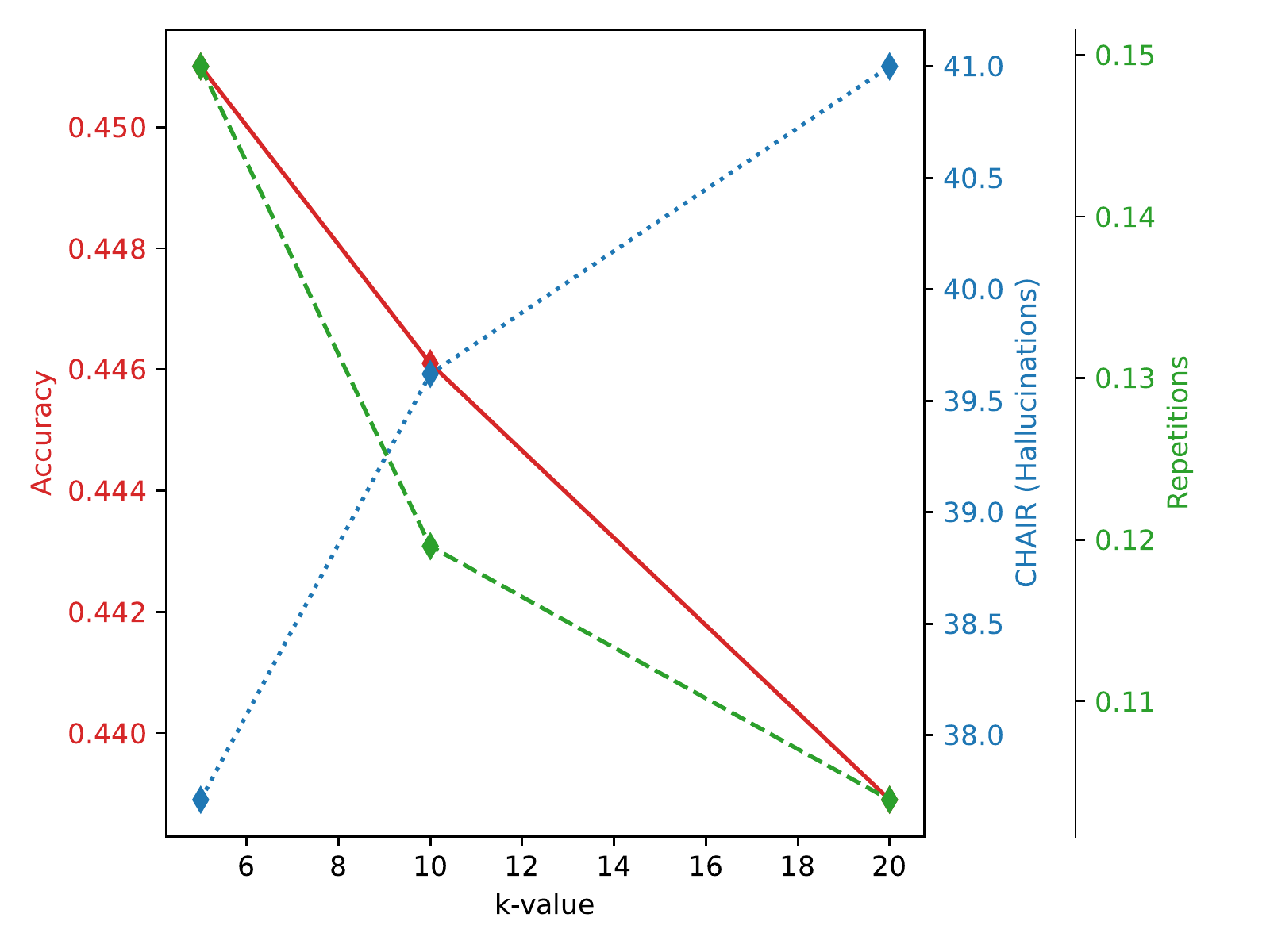} 
    \centering
    \caption{Hyper-parameter choices' effect on the accuracy, hallucinations, and repetitions in top-k sampling.}
    \label{fig:topk}
\end{figure}

\begin{figure}[h]
    \includegraphics[width=1\linewidth]{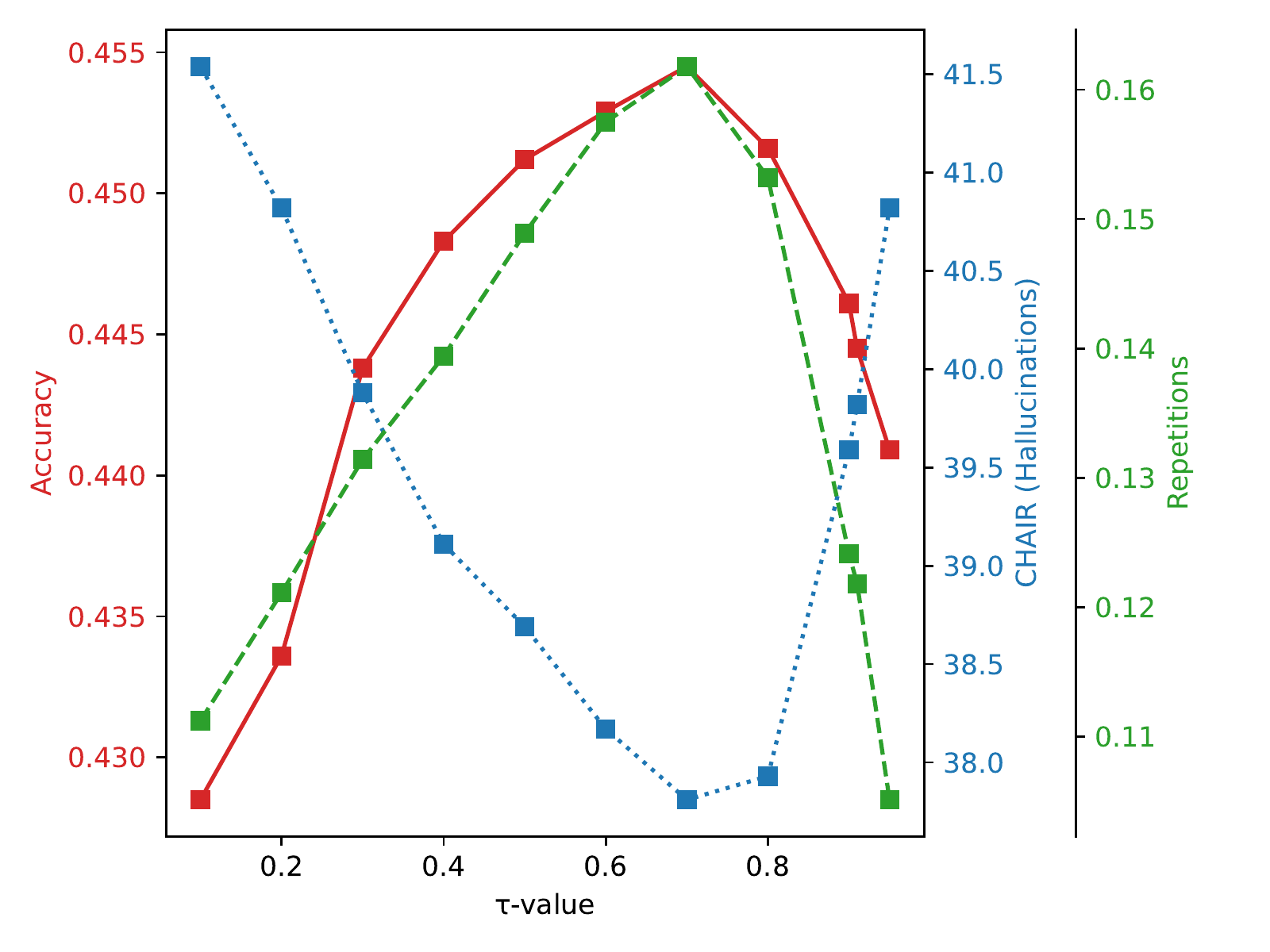} 
    \centering
    \caption{Hyper-parameter choices' effect on the accuracy, hallucinations, and repetitions in typical decoding.}
    \label{fig:typical}
\end{figure}

\begin{figure}[h]
    \includegraphics[width=1\linewidth]{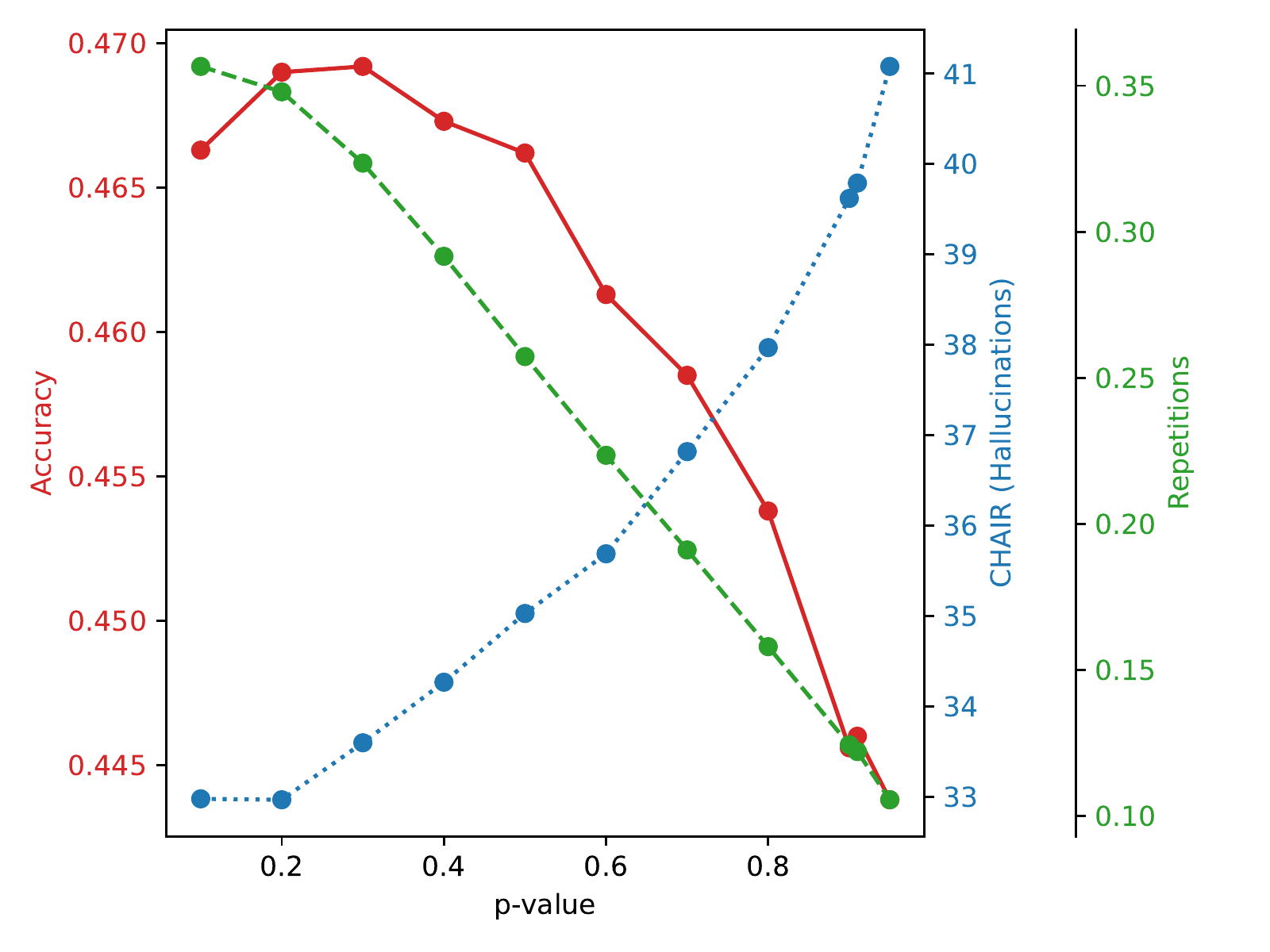} 
    \centering
    \caption{Hyper-parameter choices' effect on the accuracy, hallucinations, and repetitions in nucleus sampling.}
    \label{fig:nucleus}
\end{figure}

Figures \ref{fig:topk}, \ref{fig:typical}, and \ref{fig:nucleus} illustrate how hyper-parameter choice affects the accuracy, the hallucinations, and the repetitions. Top-k sampling (Figure \ref{fig:topk}) shows decreased accuracy and repetitions, and increased hallucinations, as the $k$-value gets higher. The same general pattern can be observed with the gradual increase of the $p$-value in nucleus sampling (Figure \ref{fig:nucleus}). On the other hand, typical decoding accuracy peaks at $\tau$ = 0.7 (Figure \ref{fig:typical}). This is also the point at which the repetitions are at their highest and the hallucinations are at their lowest. Both very high and very low $\tau$-values cause lower accuracy, fewer repetitions, and an increase of hallucinations.


\begin{table*}[]
	\centering
	\small
 	\bgroup
    \def\arraystretch{1.6}
	\begin{tabular}{|l|r|r|r|r|r|r|r|r|r|r|r|r|}
		\hline
		& \multicolumn{2}{c|}{\begin{tabular}[c]{@{}c@{}} \% Accuracy $\uparrow$ \end{tabular}} & \multicolumn{2}{c|}{CHAIR-i $\downarrow$} & \multicolumn{2}{c|}{CHAIR-s $\downarrow$} & \multicolumn{2}{c|}{\begin{tabular}[c]{@{}c@{}}\% games with\\ repetitions $\downarrow$ \end{tabular}}  & \multicolumn{2}{c|}{\begin{tabular}[c]{@{}c@{}}Vocabulary\\ Size $\uparrow$ \end{tabular}} &  \multicolumn{2}{c|}{\begin{tabular}[c]{@{}c@{}}Rare\\ Words $\uparrow$ \end{tabular}}  \\ \cline{2-13}
		& mean & std & mean & std & mean & std & mean & std & mean & std & mean & std\\ \hline
		CI & \textbf{51.39} & 0.00 & \textbf{15.09} & 0.00 & \textbf{28.48} & 0.00 & 30.33 & 0.00 & 858 & 0.0 & 34 & 0.0\\ \hline
		BS (beam = 3) & \textbf{47.05} & 0.00 & 18.33 & 0.00 & \textbf{31.08} & 0.00 & 38.49 & 0.00 & 731 & 0.0 & 27 & 0.0\\ \hline
		NS ($p$ = 0.3) & \textbf{46.92} & 0.17 & 17.96 & 0.09 & 33.60 & 0.14 & 32.35 & 0.21 & 1016 & 8.3 & 78 & 5.5\\ \hline
		NS ($p$ = 0.2) & 46.90 & 0.11 & \textbf{17.65} & 0.06 & \textbf{34.79} & 0.12 & 46.41 & 0.18 & 919 & 7.1 & 59 & 4.5\\ \hline
		NS ($p$ = 0.4) & 46.73 & 0.27 & 18.38 & 0.17 & 34.27 & 0.28 & 29.16 & 0.25 & 1097 & 11.9 & 88 & 6.4\\ \hline
		NS ($p$ = 0.1) & 46.63 & 0.02 & 17.76 & 0.01 & 32.98 & 0.01 & 35.66 & 0.06 & 839 & 1.6 & 46 & 1.9\\ \hline
		NS ($p$ = 0.5) & 46.62 & 0.17 & 19.10 & 0.09 & 35.03 & 0.10 & 25.73 & 0.26 & 1192 & 18.0 & 103 & 3.4\\ \hline
		GS & 46.58 & 0.00 & \textbf{17.75} & 0.00 & \textbf{32.97} & 0.00 & 35.63 & 0.00 & 834 & 0.0 & 46 & 0.0\\ \hline
		NS ($p$ = 0.6) & 46.13 & 0.38 & 20.04 & 0.15 & 35.69 & 0.35 & 22.35 & 0.26 & 1303 & 12.7 & 126 & 9.4\\ \hline
		NS ($p$ = 0.7) & 45.85 & 0.19 & 21.14 & 0.11 & 36.82 & 0.31 & 19.11 & 0.26 & 1451 & 9.0 & 162 & 9.5\\ \hline
		TD ($\tau$ = 0.7) & 45.45 & 0.32 & 21.84 & 0.15 & 37.81 & 0.23 & 16.18 & 0.29 & 1703 & 13.0 & 247 & 12.6\\ \hline
		NS (p = 0.8) & 45.38 & 0.14 & 22.20 & 0.16 & 37.97 & 0.30 & 15.80 & 0.19 & 1643 & 23.0 & 219 & 12.9\\ \hline
		TD ($\tau$ = 0.6) & 45.29 & 0.28 & 22.08 & 0.16 & 38.17 & 0.30 & 15.75 & 0.17 & 1723 & 21.9 & 248 & 18.3\\ \hline
		TD ($\tau$ = 0.8) & 45.16 & 0.18 & 22.21 & 0.20 & 37.93 & 0.29 & 15.32 & 0.28 & 1712 & 10.8 & 244 & 13.6\\ \hline
		TD ($\tau$ = 0.5) & 45.12 & 0.15 & 22.60 & 0.17 & 38.69 & 0.36 & 14.89 & 0.22 & 1745 & 7.3 & 262 & 8.7\\ \hline
		Top-k ($k$ = 5) & 45.10 & 0.27 & 22.84 & 0.21 & 37.71 & 0.26 & 14.93 & 0.10 & 1462 & 12.6 & 171 & 5.2\\ \hline
		TD ($\tau$ = 0.4) & 44.83 & 0.17 & 23.11 & 0.13 & 39.11 & 0.44 & 13.94 & 0.24 & 1755 & 19.0 & 265 & 12.2\\ \hline
		TD ($\tau$ = 0.9) & 44.61 & 0.16 & 23.74 & 0.18 & 39.59 & 0.19 & 12.41 & 0.25 & 1919 & 13.5 & 334 & 9.1\\ \hline
		Top-k ($k$ = 10) & 44.61 & 0.24 & 24.03 & 0.29 & 39.62 & 0.26 & 11.96 & 0.16 & 1692 & 13.8 & 235 & 10.5\\ \hline
		NS ($p$ = 0.91) & 44.60 & 0.15 & 23.92 & 0.13 & 39.79 & 0.23 & 12.21 & 0.13 & 1948 & 22.3 & 342 & 14.0\\ \hline
		NS ($p$ = 0.9) & 44.56 & 0.23 & 23.82 & 0.07 & 39.62 & 0.17 & 12.44 & 0.10 & 1912 & 20.2 & 332 & 13.6\\ \hline
		TD ($\tau$ = 0.91) & 44.45 & 0.27 & 23.92 & 0.14 & 39.82 & 0.31 & 12.18 & 0.19 & 1945 & 11.7 & 345 & 16.1\\ \hline
		TD ($\tau$ = 0.3) & 44.38 & 0.31 & 24.07 & 0.21 & 39.88 & 0.22 & 13.14 & 0.23 & 1791 & 14.8 & 278 & 13.7\\ \hline
		NS ($p$ = 0.95) & 44.38 & 0.12 & 24.93 & 0.15 & 41.08 & 0.24 & 10.56 & 0.05 & \textbf{2129} & 11.3 & \textbf{438} & 11.4\\ \hline
		TD ($\tau$ = 0.95) & 44.09 & 0.21 & 24.83 & 0.24 & 40.82 & 0.28 & \textbf{10.51} & 0.20 & \textbf{2117} & 18.9 & \textbf{435} & 17.1\\ \hline
		Top-k ($k$ = 20) & 43.89 & 0.10 & 25.12 & 0.30 & 41.00 & 0.39 & \textbf{10.39} & 0.17 & 1879 & 23.1 & 305 & 17.5\\ \hline
		TD ($\tau$ = 0.2) & 43.36 & 0.20 & 25.19 & 0.16 & 40.82 & 0.34 & 12.11 & 0.09 & 1815 & 21.5 & 287 & 10.7\\ \hline
		PS & 43.13 & 0.28 & 26.55 & 0.25 & 43.23 & 0.36 & \textbf{8.32} & 0.17 & \textbf{2609} & 9.3 & \textbf{793} & 11.4\\ \hline
		TD ($\tau$ = 0.1) & 42.85 & 0.15 & 26.25 & 0.14 & 41.54 & 0.09 & 11.12 & 0.11 & 1825 & 18.4 & 286 & 13.5\\ \hline

	\end{tabular}
	\egroup
	\caption{Comparison between decoding strategies and their hyper-parameters (CI = Confirm-it, BS = Beam Search, NS = Nucleus Sampling, GS = Greedy Search, TD = Typical Decoding, Top-k = Top-k Sampling, PS = Pure Sampling).}
	\label{tab:results_full}
\end{table*}

\subsection{Experiments}

Table \ref{tab:results_full} presents our results in detail for all the parameter configurations we considered. We have computed accuracy percentage, CHAIR-i, CHAIR-s, percentage of games with repeated questions, vocabulary size and number of rare words for each decoding method and its respective hyper-parameter configurations. These results are sorted by decreasing accuracy. The 3 best results of each metric are in bold.

\subsection{Human Annotation Details}

\begin{figure}[h]
    \includegraphics[width=1\linewidth]{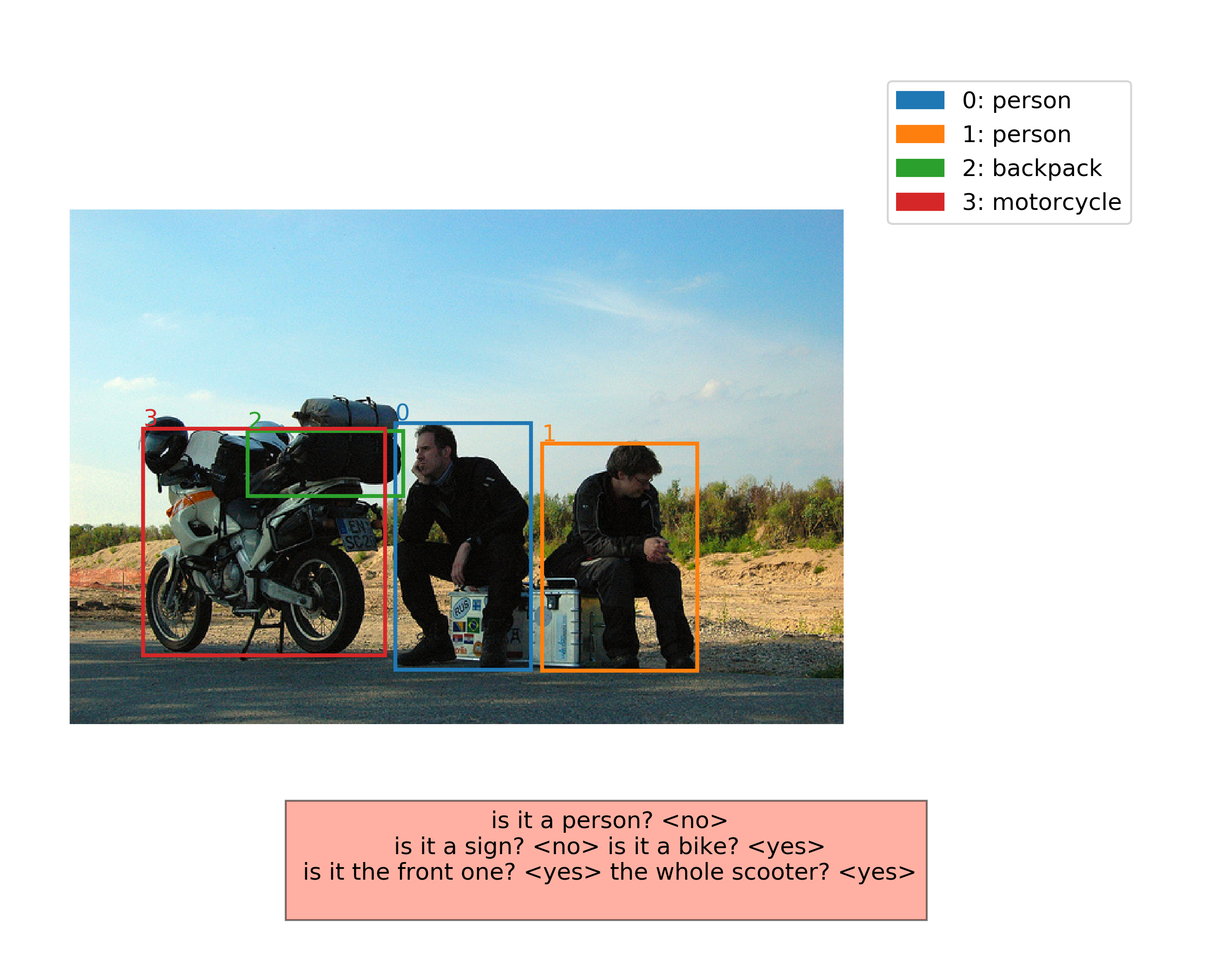} 
    \centering
    \caption{Example of the games displayed to the participants for the annotation task. Participants had to select one target object among the list of candidate objects on the right. The machine-generated dialogue is in the red box.}
    \label{fig:annotation_example}
\end{figure}

The annotation was done by 8 human annotators on a sample of GuessWhat?! games. They were recruited within our organization on a voluntary basis and they did not receive any payment for the annotation. Written informed consent was obtained from all the participants. Participants were 4 males and 4 females with high educational level and from different ethnic groups. Before the beginning of the annotation task, each annotator was briefed on the GuessWhat?! gameplay and purpose, and was asked to annotate some sample games in order to get familiar with the annotation process. We used the \url{makesense.ai} online software for image recognition. Each image had a minimum of 3 and a maximum of 6 candidate objects. The annotators could see both the bounding box and the category for each candidate object in the image. They could also see the full dialogue between the Questioner and the Oracle. The annotators then had to pick the object they believed was the right one, based on the information given by the dialogue. Figure \ref{fig:annotation_example} provides an example of the games we asked the participants to annotate. Overall, we extracted 200 images from the GuessWhat?!\@ test set and generated one dialogue per decoding strategy per image. In total, we thus generated 800 dialogues. Each human participant annotated 25 images per decoding strategy. To prevent biases, the participants were not given any information about the decoding strategy that each image represented.
 

\subsection{Comparison of Decoding Strategies}

Table \ref{tab:dialogue_examples} shows a comparison between generated dialogues of various decoding strategies (see Figure \ref{fig:birds} for the image the dialogues refer to).

\begin{figure}[h]
    \includegraphics[width=1\linewidth]{images/birds.jpg} 
    \centering
    \caption{The image for game 147344.}
    \label{fig:birds}
\end{figure}

\begin{table*}[]
	\centering
	\fontsize{8}{8}\selectfont
	\bgroup
    \def\arraystretch{1.5}
	\begin{tabular}{|c|c|c|c|c|c|c|}
		\hline
		Turn & Speaker & Confirm-it & Nucleus Sampling & Typical Decoding & Pure Sampling \\ \hline
		\multirow{2}{4em}{0} & Questioner & is it a bird? & is it a bird? & is it a bird? & is it a bird? \\ \cline{2-6}
		& Oracle & yes & yes & yes & yes \\ \hline
		\multirow{2}{4em}{1} & Questioner & is it on the left side? & is it on the left side? & is it on the air? & is it close? \\ \cline{2-6}
		 & Oracle & yes & yes & yes & yes \\ \hline
		\multirow{2}{4em}{2} & Questioner & \begin{tabular}[c]{@{}c@{}} is it on the left side\\ of the group? \end{tabular}  & \begin{tabular}[c]{@{}c@{}} is it one of the\\ first birds? \end{tabular}   & is it 1 of the 3 on the right? & is it squatting? \\ \cline{2-6}
		 & Oracle & no & yes & no & yes \\ \hline
		\multirow{2}{4em}{3} & Questioner & is it in the middle? & \begin{tabular}[c]{@{}c@{}}is it the first one \\ from the left? \end{tabular}   & \begin{tabular}[c]{@{}l@{}} is it one of the three birds \\  that are next to each other? \end{tabular}  & \begin{tabular}[c]{@{}c@{}} is it one of the three \\ all the way to the right? \end{tabular}  \\ \cline{2-6}
		 & Oracle & yes & no & yes & yes \\ \hline
		\multirow{2}{4em}{4} & Questioner & \begin{tabular}[c]{@{}c@{}}is it one of the three \\ on the right? \end{tabular}  & \begin{tabular}[c]{@{}c@{}}is it the second one \\ from the left? \end{tabular}   & is it further left? & is it 1st?  \\ \cline{2-6}
		 & Oracle & no & no & yes & no \\ \hline

	\end{tabular}
	\egroup
	\caption{Comparison between dialogues generated by various decoding strategies for the image reported in Figure \ref{fig:birds}.}
	\label{tab:dialogue_examples}
\end{table*}

\end{document}